\documentclass[twocolumn,fullpage,11pt]{article}
\usepackage{latexsym}

\newtheorem{Proposition}{Proposition}

\newcommand{\morph}{ \colon }
\newcommand{\memberof}{\,{\in}\,}
\newcommand{\define}{\stackrel{{\rm df}}{=}}
\newcommand{\tensor}{{\otimes}}
\newcommand{\implication}{{\Rightarrow}}
\newcommand{\equivalence}{{\Leftrightarrow}}
\newcommand{\minus}{\stackrel{\mbox{\bf .}}{-}}
\newcommand{\tensorimplysource}
   {\! \mbox{\hspace{.21em}--\hspace{-.21em}} \backslash }
\newcommand{\tensorimplytarget}
   { / \mbox{\hspace{-.23em}--\hspace{.23em}} \!}

\newcommand{\product}[2]{ {#1} {\times} {#2} }

\newcommand{\derivedir}[2]{ {#1}^{\prime}_{#2} }
\newcommand{\deriveinv}[2]{ {#1}^{\prime}_{#2} }

\newcommand{\pair}[2]{\langle #1,#2 \rangle}
\newcommand{\triple}[3]{\mbox{$ \langle #1,#2,#3 \rangle $}}
\newcommand{\quadruple}[4]{\mbox{$ \langle #1,#2,#3,#4 \rangle $}}
\newcommand{\quintuple}[5]{\mbox{$ \langle #1,#2,#3,#4,#5 \rangle $}}

\newcommand{\morphism}[3]{{#1} \stackrel{{#2}}{\rightarrow} {#3}}

\title{ \LARGE{\bf Enriched Interpretation}\normalsize \normalsize}

\author{
	Robert E. Kent \\
	University of Arkansas
}

\date{}

\begin{document}

\maketitle

\begin{abstract}

The theory introduced, presented and developed in this paper,
is concerned with an enriched extension
of the theory of Rough Sets pioneered by Zdzislaw Pawlak \cite{pawlak82}.
The enrichment discussed here is in the sense of valuated categories
as developed by F.W. Lawvere \cite{lawvere73}.
This paper relates Rough Sets to an abstraction 
of the theory of Fuzzy Sets pioneered by Lotfi Zadeh \cite{zadeh75},
and provides a natural foundation for {\em soft computation\/}.
To paraphrase Lotfi Zadeh,
the impetus for the transition from a hard theory to a soft theory
derives from the fact that
both the generality of a theory
and its applicability to real-world problems
are substantially enhanced
by replacing various hard concepts with their soft counterparts.
Here we discuss the corresponding enriched notions
for indiscernibility, subsets, upper/lower approximations, and rough sets.
Throughout,
we indicate linkages with the theory of Formal Concept Analysis
pioneered by Rudolf Wille \cite{wille82}.
We pay particular attention to the all-important notion of a {\em linguistic variable\/}
--- developing its enriched extension,
comparing it with the notion of conceptual scale from Formal Concept Analysis,
and
discussing the pragmatic issues of its creation and use in the interpretation of data.
These pragmatic issues are exemplified by 
the discovery, conceptual analysis, interpretation, and categorization
of networked information resources in {\sc wave},
the Web Analysis and Visualization Environment \cite{KeNe94}
currently being developed for the management and interpretation of 
the universe of resource information distributed over the World-Wide Web.
\end{abstract}

%%%%%%%%%%%%%%%%%%%%%%%%%%%%%%%%%%%%%%%%%%%%%%%%%%%%%%%%%%%%%%%%%%%%%%%%%%%%%%%%
\section{Indiscernibility}
%%%%%%%%%%%%%%%%%%%%%%%%%%%%%%%%%%%%%%%%%%%%%%%%%%%%%%%%%%%%%%%%%%%%%%%%%%%%%%%%

Indiscernibility,
a central concept in Rough Sets theory,
is traditionally treated as a {\em hard\/} relationship
--- either two objects {\em are\/} indiscernible 
or they are {\em not\/}.
In order to define and develop a {\em soft\/} theory of Rough Sets,
it would seem quite appropriate,
if not necessary,
to define and develop a soft or graded version of indiscernibility.
We do just that in this paper
by using ideas from the theory of valuated categories.

An {\em approximation space\/} \cite{pawlak82} is traditionally defined as
a pair ${\cal G} = \pair{G}{{E}}$
consisting of
a set of {\em objects\/} or {\em entities\/} $G$
and 
an equivalence relation $E \subseteq \product{G}{G}$
called {\em indiscernibility\/}.
Two objects $g_1,g_2 \memberof G$ are {\em indiscernible\/} 
when $g_1{E}g_2$;
that is,
when $E(g_1,g_2) = {\tt true}$.
Equivalently,
an {\em approximation space\/} (function version) is a triple $\triple{G}{\phi}{D}$,
where $G$ is a set of objects,
$D$ is a set (hard and unenriched!) of values,
and $G \stackrel{\phi}{\rightarrow} D$ is a (not necessarily surjective) function 
called a {\em description function\/}.
The description function $\phi$ represents a certain amount of knowledge 
about the objects in $G$.
Two objects $g_1,g_2 \memberof G$ are {\em indiscernible\/}
when 
the procedure $\phi$ cannot distinguish between them,
$\phi(g_1) = \phi(g_2)$;
or more generally,
when ${\rm Eq}_D(\phi(g_1),\phi(g_2)) = {\tt true}$
for some sense and relationship ${\rm Eq}_D$ 
of identification or approximation of values in $D$.
We are particularly interested in the case where
$D = \wp M \cong 2^M$ consists of subsets of a collection of attributes $M$,
and $\phi$ maps an object of $G$ to the subset of all attributes that it satisfies.

One way to soften this definition
is to observe the fact
that often $D$ has additional enriched structure
--- either order-theoretic, topological or algebraic structure.
To ignore this structure 
is to weaken the Rough Set analysis of the situation
by using only the less refined, harder representation.
In this paper we develop a more general, more flexible, and softer approach 
to Rough Sets,
which handles enriched order-theoretic, metric topological, and fuzzy structure.
A full categorical formulation would also handle algebraic structure.
To enrich
(and yes, fuzzify)
Rough Set notions,
we allow grades of indiscernibleness
by assuming that $D$ has {\bf V}-enriched structure on it,
where {\bf V} = $\quintuple{V}{\preceq}{\tensor}{\implication}{e}$
is a closed preorder (see Appendix~\ref{closed_preorder});
that is,
we assume that $D$ is an approximation {\bf V}-space.

%%%%%%%%%%%%%%%%%%%%%%%%%%%%%%%%%%%%%%%%%%%%%%%%%%%%%%%%%%%%%%%%%%%%%%%%%%%%%%%%
\section{Spaces and Maps}
%%%%%%%%%%%%%%%%%%%%%%%%%%%%%%%%%%%%%%%%%%%%%%%%%%%%%%%%%%%%%%%%%%%%%%%%%%%%%%%%

While enriched approximation spaces are the appropriate abstraction of indiscernibility
and our main concern in this paper,
it seems that these approximation spaces are best defined in terms of 
an asymmetric generalization called simply an enriched space.
A pair ${\cal X} = \pair{X}{\mu}$ consisting of 
a set $X$ and a function $\mu \morph \product{X}{X} \rightarrow {\bf V}$ 
is called a {\em {\bf V}-enriched space\/} or {\em {\bf V}-space}
when it satisfies 
\begin{description}
	\item[reflexivity (zero law):] 
		$e \preceq \mu(x,x)$,
		for all $x \memberof X$;
	\item[transitivity (triangle axiom):] 
		$\mu(x_1,x_2) \tensor \mu(x_2,x_3) \preceq \mu(x_1,x_3)$,
		for all $x_1,x_2,x_3 \memberof X$.
\end{description}
The function $\mu$, called a {\em metric\/},
represents a measure of agreement or distance between the elements of $X$.
We view the metric $\mu$ to be a special square matrix 
$\mu = \left[ \mu_{x_i,x_j} \right]$ of {\bf V}-values.
We can interpret $\mu$ to be 
either 
an enriched preordering,
a generalized distance function,
a similarity measure
or 
a gradation.

When {\bf V} = {\bf 2}, the Boolean case,
a {\bf V}-space ${\cal X}$ is precisely a preorder 
${\cal X} = \pair{X}{\preceq}$
with order characteristic function
${\preceq} \morph \product{X}{X} \rightarrow {\bf 2}$.
When {\bf V} = \boldmath$\Re$\unboldmath, the metric topology case,
a {\bf V}-space ${\cal X}$ is (generalize) metric space 
${\cal X} = \pair{X}{\delta}$
with distance function
$\delta \morph \product{X}{X} \rightarrow \mbox{\boldmath$\Re$\unboldmath}$.
When {\bf V} = {\bf [0,1]}, the fuzzy case,
a {\bf V}-space ${\cal X}$ is a fuzzy space 
${\cal X} = \pair{X}{\mu}$
with similarity measure
$\mu \morph \product{X}{X} \rightarrow {\bf [0,1]}$.
Any {\bf V}-space ${\cal X} = \pair{X}{\mu}$ has 
a {\em dual\/} or {\em opposite\/} {\bf V}-space 
${\cal X}^{\rm op} = \pair{X}{\mu^{\rm op}}$,
where $\mu^{\rm op}(x_1,x_2) = \mu(x_2,x_1)$ is the dual or opposite metric.
The {\em sum\/} of any two {\bf V}-spaces 
${\cal X}_0 = \pair{X_0}{\mu_0}$ and ${\cal X}_1 = \pair{X_1}{\mu_1}$
is the {\bf V}-space 
${\cal X}_0 \oplus {\cal X}_1 = \pair{X_0{+}X_1}{\mu}$ 
defined by
$\mu(x_0,x_0^\prime) = \mu_0(x_0,x_0^\prime)$,
$\mu(x_0,x_1)        = \bot_{\rm V}$,
$\mu(x_1,x_0)        = \bot_{\rm V}$, and
$\mu(x_1,x_1^\prime) = \mu_1(x_1,x_1^\prime)$.
In general our metrics are asymmetrical:
$\mu(x_1,x_2) \neq \mu(x_2,x_1)$.
A {\em {\bf V}-enriched approximation space\/} 
or {\em approximation {\bf V}-space\/}
is defined to be a symmetrical {\bf V}-space.
So the metric $\mu$,
called an {\em indiscernibility measure\/},
is a {\bf V}-enriched equivalence relation on $X$ 
satisfying reflexivity, transitivity and
\begin{description}
	\item[symmetry:] 
		$\mu(x_2,x_1) = \mu(x_1,x_2)$,
		for all $x_1,x_2 \memberof X$.
\end{description}
Any {\bf V}-space ${\cal X} = \pair{X}{\mu}$ can be symmetrized 
and made into an approximation space,
by defining the {\em junction metric\/}
$\mu^{\rm sym}(x_1,x_2) 
	=
	\mu(x_1,x_2) \tensor \mu^{\rm op}(x_1,x_2)$.

Associated with every {\bf V}-space ${\cal X} = \pair{X}{\mu}$
is an {\em underlying preorder\/} 
$\Box_{\rm V}({\cal X}) = \pair{X}{\preceq}$
where $x_1 \preceq x_2$ when $e \preceq \mu(x_1,x_2)$,
and $x_1$ and $x_2$ are unrelated when $e \not\preceq \mu(x_1,x_2)$.
Two elements $x_1,x_2 \memberof X$ are said to be {\em indiscernible\/} 
when $x_1 \equiv x_2$,
where $x_1 \equiv x_2$ means $x_1 \preceq x_2$ and $x_2 \preceq x_1$.
A {\bf V}-space ${\cal X} = \pair{X}{\mu}$ is {\em strict\/}
when 
the underlying indiscernibility relation is the identity:
if $x_1 \equiv x_2$ then $x_1 = x_2$.
The set of ``truth values with implication''
${\cal V} = \pair{V}{\implication}$ is a strict {\bf V}-space.
Note that 
$\Box({\cal X}^{\rm op}) 
	= (\Box{\cal X})^{\rm op} = \pair{X}{\succeq}$
the opposite order,
and that 
$\Box_{\rm V}({\cal X}^{\rm sym}) 
	= (\Box_{\rm V}({\cal X}))^{\rm sym} 
	= \pair{X}{\equiv}$
the underlying indiscernibility relation.
For a strict {\bf V}-space 
the underlying preorder is a partial order.
For a (soft) approximation {\bf V}-space
the underlying preorder is a (hard) equivalence relation.
For the space of generalized truth values ${\cal V} = \pair{V}{\implication}$,
since $e \preceq v_1 \implication v_2$ iff $v_1 \preceq v_2$,
the underlying preorder is the given order on ${\bf V}$.

A {\bf V}-{\em map} 
$f \morph {\cal X} \rightarrow {\cal Y}$
between two {\bf V}-spaces 
${\cal X} = \pair{X}{\mu}$ and ${\cal Y} = \pair{Y}{\nu}$
is a function $f \morph X \rightarrow Y$ 
that preserves measure
by satisfying the condition
$\mu(x_1,x_2) \preceq \nu(f(x_1),f(x_2))$ for all $x_1,x_2 \in X$.
When this condition is an equality at all elements
$\mu(x_1,x_2) = \nu(f(x_1),f(x_2))$ for all $x_1,x_2 \in X$,
$f \morph {\cal X} \rightarrow {\cal Y}$
is called a {\em {\bf V}-isometry\/}.
Any function $f \morph X \rightarrow {\cal Y}$
from a set $X$ into a space ${\cal Y}$
induces a {\bf V}-space metric $\mu_f$ on the set $X$,
defined by $\mu_f(x_1,x_2) = \nu(f(x_1),f(x_2))$,
and making
$f \morph \pair{X}{\mu_f} \rightarrow {\cal Y}$
an isometry.
By modus ponens,
$(\:) \tensor v \morph {\cal V} \rightarrow {\cal V}$
is a {\bf V}-map for all elements $v \memberof V$.
By transitivity of implication,
$v \implication (\:) \morph {\cal V} \rightarrow {\cal V}$
is a {\bf V}-map for all elements $v \memberof V$.
When {\bf V} = {\bf 2}, the Boolean case,
a {\bf V}-map $f \morph {\cal X} \rightarrow {\cal Y}$ 
is precisely a monotonic function.
When {\bf V} = \boldmath$\Re$\unboldmath, the metric topology case,
a {\bf V}-map $f \morph {\cal X} \rightarrow {\cal Y}$ 
is precisely a contraction.
When {\bf V} = {\bf [0,1]}, the fuzzy case,
a {\bf V}-map $f \morph {\cal X} \rightarrow {\cal Y}$ 
is a fuzzy measure preserving function.
Each {\bf V}-map $\morphism{{\cal X}}{f}{{\cal Y}}$
is a monotonic function between the underlying ordered sets.
Two {\bf V}-maps in opposite directions
$\morphism{{\cal X}}{f}{{\cal Y}}$
and
$\morphism{{\cal Y}}{g}{{\cal X}}$
form an {\em enriched adjointness (Galois connection)\/} 
denoted by $f \dashv g$ 
when
\[ \mu_Y(f(x),y) = \mu_X(x,g(y)) \]
for all $x \in X$ and $y \in Y$.
Two such maps then form an adjointness as monotonic functions.
{\bf V}-spaces and {\bf V}-maps form the category ${\bf Space}_{\rm V}$
with obvious underlying functor
${\bf Space}_{\rm V} \stackrel{\Box_{\rm V}}{\rightarrow} {\bf Space}_{\rm 2}$,
where
${\bf Space}_{\rm 2}$ is the category of preorders and monotonic functions.

%%%%%%%%%%%%%%%%%%%%%%%%%%%%%%%%%%%%%%%%%%%%%%%%%%%%%%%%%%%%%%%%%%%%%%%%%%%%%%%%
\section{Relations}
%%%%%%%%%%%%%%%%%%%%%%%%%%%%%%%%%%%%%%%%%%%%%%%%%%%%%%%%%%%%%%%%%%%%%%%%%%%%%%%%

Each element $x \memberof X$ of a {\bf V}-space ${\cal X} = \pair{X}{\mu}$
can be represented as the {\bf V}-predicate ${\rm y}(x) = \mu(x,-)$ over ${\cal X}$
where ${\rm y}(x)(x') = \mu(x,x')$ for each element $x' \memberof X$.
The function ${\rm y}_X \morph X \rightarrow V^X$,
which is called the {\em Yoneda embedding},
is a {\bf V}-isometry 
${\rm y}_{\cal X} \morph {\cal X}^{\rm op} \rightarrow {\bf V}^{\cal X}$.
Composition of (the opposite of) a {\bf V}-map
$f \morph {\cal X} \rightarrow {\cal Y}$
on the right with the Yoneda embedding
${\rm y}_{\cal Y} \morph {\cal Y}^{\rm op} \rightarrow {\bf V}^{\cal Y}$,
resulting in the {\bf V}-map
$f_\ast \morph {\cal X}^{\rm op} \rightarrow {\bf V}^{\cal Y}$,
allows us to generalize the concept of a {\bf V}-map.
Such a generalized {\bf V}-map,
equivalent to a {\bf V}-map
$\product{{\cal X}^{\rm op}}{{\cal Y}} \stackrel{\tau}{\longrightarrow} {\bf V}$,
may be regarded to be 
a {\em {\bf V}-enriched relation\/} or {\em {\bf V}-relation\/} 
from {\cal X} to {\cal Y}.
It is denoted by ${\cal X} \stackrel{\tau}{\rightharpoondown} {\cal Y}$,
with $\tau(x,y)$ an element of {\bf V} interpreted as the
``truth-value of the $\tau$-relatedness of $x$ to $y$'' \cite{lawvere73}.
A {\bf V}-relation is an
$\product{|{\cal X}|}{|{\cal Y}|}$-matrix,
whose $(x,y)$-th entry is $\tau(x,y)$.
In elementary terms,
a {\bf V}-relation is an
$\product{|X|}{|Y|}$-matrix,
which respects the measures on both left and right:
$\mu(x^\prime,x) \tensor \tau(x,y) \preceq \tau(x^\prime,y)$
and
$\tau(x,y) \tensor \nu(y,y^\prime) \preceq \tau(x,y^\prime)$,
$\forall x,x^\prime \in X, y,y^\prime \in Y$.

As mentioned above,
every {\bf V}-map 
${\cal X} \stackrel{f}{\rightarrow} {\cal Y}$
determines a {\bf V}-relation 
${\cal X} \stackrel{f_\ast}{\rightharpoondown} {\cal Y}$
defined by $f_\ast = f^{\rm op} \cdot {\rm y}_{\cal Y}$,
or on elements by $f_\ast(x,y) = \nu(f(x),y)$.
In particular,
the Yoneda embedding becomes the relation ${\cal X} \stackrel{\mu}{\rightharpoondown} {\cal X}$.
Dually every {\bf V}-map 
${\cal X} \stackrel{f}{\rightarrow} {\cal Y}$ 
also determines a {\bf V}-relation 
${\cal Y} \stackrel{f^\ast}{\rightharpoondown} {\cal X}$
in the opposite direction
defined by $f^\ast = {\rm y}_{\cal Y} \cdot {\bf V}^f$,
or on elements by $f^\ast(y,x) = \nu(y,f(x))$.

A pair of {\bf V}-relations 
${\cal X} \stackrel{\sigma}{\rightharpoondown} {\cal Y}$ 
and 
${\cal Y} \stackrel{\tau}{\rightharpoondown} {\cal Z}$
can be composed,
yielding the {\bf V}-relation 
${\cal X} \stackrel{\sigma\circ\tau}{\rightharpoondown} {\cal Z}$
defined to be the supremum (iterated disjunction)
\[
	(\sigma \circ \tau)(x,z)
%	= \int^{y \in {\cal Y}} \left( \sigma(x,y) \tensor \tau(y,z) \right)
	= \bigvee_{y \in {\cal Y}} \left( \sigma(x,y) \tensor \rho(y,z) \right)
\]
Relational composition is viewed as matrix multiplication.
One can verify that relational composition is associative
$(\rho \circ \sigma) \circ \tau = \rho \circ (\sigma \circ \tau)$,
and that metrics (as {\bf V}-relations) are identities
$\mu \circ \tau = \tau = \tau \circ \nu$.
So {\bf V}-spaces and {\bf V}-relations form a category ${\bf Rel}_{\rm V}$.
One can also verify that
$(f \cdot g)_\ast = f_\ast \circ g_\ast$
for any two composable {\bf V}-maps 
${\cal X} \stackrel{f}{\rightarrow} {\cal Y} \stackrel{g}{\rightarrow} {\cal Z}$,
and that 
$(\mbox{Id}_{\cal X})_\ast = \mu$ 
the identity {\bf V}-relation at ${\cal X}$.
So the Yoneda embedding determines a functor
$(\:)_\ast \morph {\bf Space}_{\rm V} \rightarrow {\bf Rel}_{\rm V}$
which makes concrete the concept generalization discussed at the beginning of this section.

Relational composition has a right adjoint called residuation.
The {\em residuation\/} of a pair of {\bf V}-relations 
${\cal X} \stackrel{\sigma}{\rightharpoondown} {\cal Y}$ 
and 
${\cal X} \stackrel{\rho}{\rightharpoondown} {\cal Z}$,
denoted by the {\bf V}-relation 
${\cal Y} \stackrel{\sigma\tensorimplysource\rho}{\rightharpoondown} {\cal Z}$,
is defined to be the infimum (iterated conjunction)
\[
	(\sigma \tensorimplysource \rho)(y,z)
%	= \int_{x \in {\cal X}} \left( \sigma(x,y) \implication \rho(x,z) \right)
	= \bigwedge_{x \in {\cal X}} \left( \sigma(x,y) \implication \rho(x,z) \right)
\]
Note that
$(\sigma \circ \tau) \tensorimplysource \rho
	= (\tau \tensorimplysource (\sigma \tensorimplysource \rho))$
for any pair of composable {\bf V}-relations 
${\cal X} \stackrel{\sigma}{\rightharpoondown} {\cal Y}$ 
and 
${\cal Y} \stackrel{\tau}{\rightharpoondown} {\cal W}$,
and that
$\mu \tensorimplysource \rho = \rho$
for identity relation ${\cal X} \stackrel{\mu}{\rightharpoondown} {\cal X}$.

%%%%%%%%%%%%%%%%%%%%%%%%%%%%%%%%%%%%%%%%%%%%%%%%%%%%%%%%%%%%%%%%%%%%%%%%%%%%%%%%
\section{Subsets}
%%%%%%%%%%%%%%%%%%%%%%%%%%%%%%%%%%%%%%%%%%%%%%%%%%%%%%%%%%%%%%%%%%%%%%%%%%%%%%%%

Given any two {\bf V}-spaces 
${\cal X} = \pair{X}{\mu}$ and ${\cal Y} = \pair{Y}{\nu}$ 
there is a {\bf V}-space ${\cal X} \tensor {\cal Y}$,
called the {\em tensor product\/} of ${\cal X}$ and ${\cal Y}$,
which enriches the Cartesian product set $\product{X}{Y}$
with the metric defined by
\[ (\mu \tensor \nu)((x_1,y_1),(x_2,y_2)) = \mu(x_1,x_2) \tensor \nu(y_1,y_2) \]
When {\bf V} is Cartesian closed,
the tensor product is the (ordinary) Cartesian product.
This tensor product construction has a right adjoint exponential construction
making ${\bf Space}_{\rm V}$ into a closed category \cite{lawvere73}.
Given any two {\bf V}-spaces ${\cal X}$ and ${\cal Y}$ the set
of all {\bf V}-maps from ${\cal X}$ to ${\cal Y}$ is a {\bf V}-space
${\cal Y}^{\cal X}$,
called the {\em exponential {\bf V}-space\/} of $\cal X$ and ${\cal Y}$,
whose {\em pointwise inf metric\/} $\mu$ is defined by
$\mu(f,g) = \bigwedge_{x \in X} \mu_Y(f(x),g(x))$.
Notice that the metric $\mu_X$ is not used to define $\mu$.
The metric $\mu_X$ is only used to restrict admission to the underlying set of ${\cal Y}^{\cal X}$.

As an important special case,
the {\em power {\bf V}-space\/} ${\bf V}^{{\cal X}}$
of all {\bf V}-valued {\bf V}-maps on ${\cal X}$
is a {\bf V}-space with metric
\[ \phi \implication \psi
= \bigwedge_{x \in X} \left( \phi(x) \implication \psi(x) \right) \]
We interpret an element of ${\bf V}^{\cal X}$,
a {\bf V}-map $\phi \morph {\cal X} \longrightarrow {\bf V}$,
to be a {\bf V}-enriched subset,
which satisfies the internal pointwise metric constraint $\mu$:
$\mu(x_1,x_2) \preceq \phi(x_1) \implication \phi(x_2)$ for all $x_1,x_2 \in X$;
or equivalently,
by the $\tensor$-$\implication$ adjointness,
$\phi(x_1) \tensor \mu(x_1,x_2) \preceq \phi(x_2)$ for all $x_1,x_2 \memberof X$.
Such a characteristic function
$\phi \morph {\cal X} \rightarrow {\bf V}$,
which is constrained by the metric on ${\cal X}$,
is called 
a {\em {\bf V}-enriched predicate\/} or {\em {\bf V}-predicate\/} in ${\cal X}$.
It can also be called,
using Rough Set terminology,
a {\em {\bf V}-enriched definable subset\/} or {\em {\bf V}-definable subset\/} in ${\cal X}$.
To use a slogan,
``predicate (or definable subset) $\equiv$ metric-constrained character''.
For the power space ${\bf V}^{\cal X}$ of {\bf V}-predicates over ${\cal X}$
the underlying preorder is the usual {\em entailment order} on {\bf V}-predicates over ${\cal X}$,
defined by $\phi(x) \preceq \psi(x)$ for all $x \memberof X$.
Associated with any {\bf V}-predicate $\phi \morph {\cal X} \rightarrow {\bf V}$
is an ordinary subset $\{{\cal X} \mid \phi\} \subseteq X$,
called the {\em extension\/} of $\phi$,
and defined by
$\{{\cal X} \mid \phi\} = \{ x \in X \mid e \preceq \phi(x) \}$.

%%%%%%%%%%%%%%%%%%%%%%%%%%%%%%%%%%%%%%%%%%%%%%%%%%%%%%%%%%%%%%%%%%%%%%%%%%%%%%%%
\section{Enriched Concept Analysis}
%%%%%%%%%%%%%%%%%%%%%%%%%%%%%%%%%%%%%%%%%%%%%%%%%%%%%%%%%%%%%%%%%%%%%%%%%%%%%%%%

Enriched Concept Analysis starts with the primitive notion of 
an enriched formal context.
A {\em (formal {\bf V}-context\/} is a triple $\triple{{\cal G}}{{\cal M}}{\iota}$
consisting of two approximation spaces 
${\cal G} = \pair{G}{\gamma}$ and ${\cal M} = \pair{M}{\mu}$
and an {\em incidence\/} {\bf V}-relation 
${\cal G} \stackrel{\iota}{\rightharpoondown} {\cal M}$
between ${\cal G}$ and ${\cal M}$.
Intuitively, 
the elements of $G$ are thought of as {\em entities\/} or {\em objects\/}
with (a priori) approximation structure $\gamma$ on objects,
the elements of $M$ are thought of as {\em properties\/}, {\em characteristics\/} or {\em attributes\/}
with approximation structure $\mu$ on attributes,
and $\iota(g,m) = v$ asserts that ``object $g$ {\em has\/} attribute $m$ with measure $v$.''

Enriched Formal Concept Analysis is based upon the understanding that
an enriched concept is a unit of thought consisting of two parts:
its extension and its intension.
Within the restricted scope of a formal context,
the {\em extent\/} of a concept is an enriched subset of objects
$\phi \in {\bf V}^{\cal G}$ 
consisting of all objects belonging to the concept,
whereas
the {\em intent\/} of a concept is a enriched subset of attributes 
$\psi \in {\bf V}^{\cal M}$ 
which includes all attributes shared by the objects.
A concept of a given context will consist of an extent/intent pair
\[ (\phi,\psi). \]
Of central importance in concept construction are two {\em derivation operators\/}
which define the notion of ``sharing'' or ``commonality''.
For any subset of objects $\phi \in {\bf V}^{{\cal G}^{\rm op}} = {\bf V}^{\cal G}$,
regarded as a {\bf V}-relation 
${\cal G} \stackrel{\phi}{\rightharpoondown} {\bf 1}$,
the direct derivation along $\iota$ is defined to be
$ \derivedir{\phi}{\iota} = \phi \tensorimplysource \iota ,$
which pointwise is
$\derivedir{\phi}{\iota}(m) 
	=  \bigwedge_{g \in G} \left( \phi(g) \implication \iota(g,m) \right)$,
the {\bf V}-subset of $M$ which for each attribute $m \in M$
provides a soft measurement of the degree
to which $m$ is an attribute of all objects in $\phi$.
For any subset of attributes $\psi \in {\bf V}^{\cal M}$
regarded as a {\bf V}-relation 
${\bf 1} \stackrel{\psi}{\rightharpoondown} {\cal M}$,
the inverse derivation along $\iota$ is defined to be
$ \deriveinv{\psi}{\iota} = \iota \tensorimplytarget \psi ,$
which pointwise is
$\deriveinv{\psi}{\iota}(g)
	=  \bigwedge_{m \in M} \left( \psi(m) \implication \iota(g,m) \right)$,
the {\bf V}-subset of $G$ which for each object $g \in G$
provides a soft measurement of the degree 
to which $g$ has all attributes in $\psi$.
These two derivation operators form an enriched adjointness
\[ \derivedir{\phi}{\iota} \Leftarrow \psi \;=\; \phi \implication \deriveinv{\psi}{\iota}. \]

To demand that a concept $(\phi,\psi)$ be determined {\em softly\/} by its extent and its intent
means that this adjointness should be
a soft inverse relationship at the extent/intent pair $(\phi,\psi)$:
the intent should contain approximately (with measure the truthvalue $v$) 
those attributes shared by all objects in the extent
$v \preceq \bigwedge_{m \in M} \left( \derivedir{\phi}{\iota}(m) \equivalence \psi(m) \right)$,
and vice-versa,
the extent should contain approximately 
those objects sharing all attributes in the intent
$v \preceq \bigwedge_{g \in G} \left( \phi(g) \implication \deriveinv{\psi}{\iota}(g) \right)$.
Together this means that
$v \preceq \left( \derivedir{\phi}{\iota} \equivalence \psi \right) \tensor \left( \phi \implication \deriveinv{\psi}{\iota} \right)$.
A {\em hard concept\/} $(\phi,\psi)$
is a concept whose extent and intent determine each other exactly,
satisfying the condition
\[  e \preceq \left( \derivedir{\phi}{\iota} \equivalence \psi \right) \tensor \left( \phi \implication \deriveinv{\psi}{\iota} \right) . \]
The collection of all hard concepts is enriched
by a generalization-specialization metric.
One concept $(\phi_1,\psi_1)$ 
is more specialized (and less general) than 
another concept $(\phi_2,\psi_2)$ with measure
$\phi_1 \implication \phi_2 = \bigwedge_{g \in G} \left( \phi_1(g) \implication \phi_2(g) \right)$;
or equivalently,
$\psi_2 \implication \psi_1 = \bigwedge_{m \in M} \left( \psi_2(m) \implication \psi_1(m) \right)$.
Concepts with this generalization-specialization metric
form a concept hierarchy for the context.
\begin{Proposition}
	The concept hierarchy is a complete {\bf V}-space 
	${\cal B}\triple{\cal G}{\cal M}{\iota}$
	called the enriched concept lattice of $\triple{\cal G}{\cal M}{\iota}$.
\end{Proposition}
Completeness means that the underlying preorder
$\Box({\cal B}\triple{\cal G}{\cal M}{\iota})$
is a complete lattice.
The join of a collection of concepts 
represents the common attributes 
(common properties, or shared characteristics) of the concepts.
The bottom of the conceptual hierarchy (the empty join)
represents the most specific concept whose intent consists of all attributes.
The meet of a collection of concepts
represents a coordinated sum of the attributes of the concepts.
The top of the conceptual hierarchy (the empty meet)
represents the universal concept whose extent consists of all objects.

According to Formal Concept Analysis,
in the formal context $\triple{G}{M}{I}$
an implication $Y_1 \rightarrow Y_2$
holds between a pair of attribute subsets $Y_1,Y_2 \subseteq M$
when
each object from $G$ having all attributes of $Y_1$ 
has also all attributes of $Y_2$.
The set of all implications forms a preorder ({\bf 2}-space).

We here define a softer notion of implication in enriched formal contexts.
For any pair of attribute predicates $\psi_1, \psi_2 \in {\bf V}^{\cal M}$,
a {\em witness\/} for the potential intuitive implication $\psi_1 \rightarrow \psi_2$
is an object $g \memberof G$ which satisfies the condition
``if $g$ has all attributes of $\psi_1$ 
then $g$ also has all attributes of $\psi_2$''.
Witnesses help verify potential implications by their collective measurement.
We collect together all witnesses for the potential implication $\psi_1 \rightarrow \psi_2$,
and we measure the ``implication witness'' 
by using the metric for {\bf V}-predicates over approximation space ${\cal G}$.
\[	\psi_1 \rightarrow \psi_2 = \deriveinv{\psi_1}{\iota} \implication \deriveinv{\psi_2}{\iota}	 \]
Let ${\bf Impl}(\triple{\cal G}{\cal M}{\iota}) = \pair{{\bf V}^{\cal M}}{\rightarrow}$ 
denote the {\bf V}-space of implications of $\triple{G}{M}{I}$.
We can, of course, limit implication pairs by requiring a certain threshhold measure
$v \preceq \psi_1 \rightarrow \psi_2$.
The notion of implication from Formal Concept Analysis is 
the derived notion of maximal implication,
requiring maximal measure
$e \preceq \psi_1 \rightarrow \psi_2$.
These implications are orderings in the underlying preorder
$\Box({\bf Impl}(\triple{\cal G}{\cal M}{\iota}))$.

%Since $Y_1 \rightarrow Y_2$ iff $Y_1 \rightarrow m$ for all attributes $m \memberof Y_2$,
%the implication $Y_1 \rightarrow Y_2$ holds iff 
%the lattice meet of the collection of concepts generated by attributes in $Y_1$
%is a specialization of
%each concept generated by an attribute $m \memberof Y_2$.
%This allows implications to be easily visualized in the concept lattice line diagram.

%%%%%%%%%%%%%%%%%%%%%%%%%%%%%%%%%%%%%%%%%%%%%%%%%%%%%%%%%%%%%%%%%%%%%%%%%%%%%%%%
\section{Linguistic Variables}
%%%%%%%%%%%%%%%%%%%%%%%%%%%%%%%%%%%%%%%%%%%%%%%%%%%%%%%%%%%%%%%%%%%%%%%%%%%%%%%%

We describe linguistic variables in terms of a use-case scenario.
We start with a collection of objects ${\cal G} = \pair{G}{\gamma}$.
We assume that some observations or experimental measurements have been made,
resulting in the production of some ``raw'' data ${\cal D} = \pair{D}{\delta}$.
Both objects and data have been enriched as approximation spaces
for benefit of flexibility by using soft structures.
The data is associated with the objects by a map
called a {\em description function\/}
\[	{\cal G} \stackrel{\phi}{\rightarrow} {\cal D}.	\]

We will use linguistic variables in order 
(1) to interpret this data 
and 
(2) to provide a view or facet of it which is meaningful to the user.
The creation of linguistic variables is an act of interpretation.
Mathematically,
the notion of a linguistic value (or constraint) is represented here 
by the notion of an enriched subset.
A {\em linguistic value\/} 
over data domain ${\cal D} = \pair{D}{\delta}$
is an enriched subset in ${\bf V}^{\cal D}$.
A {\em linguistic variable\/} \cite{zadeh75}
({\em conceptual scale\/} \cite{ganter89}
or {\em distributed constraint\/} \cite{kent93b})
over data domain ${\cal D} = \pair{D}{\delta}$
is a collection
$\sigma = \{ \sigma_m \in {\bf V}^{\cal D} \mid m \in M \}$
of linguistic values over ${\cal D}$,
indexed by a collection of attribute symbols $M$.
Using functional notation
we can write this as the {\bf V}-map
$\sigma \morph {\cal M} \rightarrow {\bf V}^{\cal D}$,
where we have enriched the attributes to a (approximation) space ${\cal M} = \pair{M}{\mu}$.
Equivalently,
a linguistic variable can be represented 
either as the map
$\tilde{\sigma} \morph {\cal D} \rightarrow {\bf V}^{\cal M}$
where
$\tilde{\sigma}(d)(m) = \sigma(m)(d)$
or as the relation
\[ {\cal M} \stackrel{\sigma}{\rightharpoondown} {\cal D} \]
where
$\sigma(m,d) = \sigma(m)(d)$.
The four parts of a linguistic variable can be interpreted as follows.
\begin{enumerate}
	\item ${\cal D}$ gives its (raw) data scope or range,
	\item ${\bf V}$ represents our interpretation bias or style,
	\item ${\cal M}$ gives linguistic terms of the linguistic variable
		which are meaningful to us,
		with a priori (approximation) measure.
	\item $\sigma$ connects, attaches or assigns (as you will)
		linguistic values to linguistic terms.
\end{enumerate}
These are listed in order of volatility
--- of these four,
${\cal D}$ varies slowest (it is given to us),
whereas
$\sigma$ is most volatile.
A standard example of a linguistic variable is ``age'',
where
\footnotesize
\begin{center}
$\begin{array}{r@{\hspace{5mm}=\hspace{5mm}}l}
	{\cal D}	&	\{0,1,\ldots,100\}	\\
	{\bf V}	&	\mbox{the Fuzzy closed poset}	\\
	{\cal M}	&	\{\mbox{``young''},\mbox{``middle-age''},\mbox{``old''}\}	\\
	\sigma(\mbox{``young''})(d)	
			&	\left\{ \begin{array}{cl}
						1,				&	0  \leq d \leq 20	\\
						-\frac{1}{20}d+2,	&	20 \leq d \leq 40	\\
						0,				&	40 \leq d
				\end{array} \right.	\\
	\multicolumn{1}{c}{\mbox{etc.}}
\end{array}$
\end{center}
\normalsize
There are two ways to combine linguistic variables,
through summation and tensoring.
\begin{description}
	\item[Constraint Sum:]
		Given two linguistic variables on the same data domain
		${\cal M}_0 \stackrel{\sigma_0}{\rightharpoondown} {\cal D}$
		and
		${\cal M}_1 \stackrel{\sigma_1}{\rightharpoondown} {\cal D}$,
		the {\em copairing\/} is the linguistic variable
		\[ {\cal M}_0 \oplus {\cal M}_1 \stackrel{[\sigma_0,\sigma_1]}{\rightharpoondown} {\cal D} \]
		on the unconstrained (or constrained) sum space of terms,
		which sums the term assignments
		\footnotesize
		\begin{center}\begin{tabular}{l}
			$[\sigma_0,\sigma_1](m_0,d) = \sigma_0(m_0,d)$ \\
			$[\sigma_0,\sigma_1](m_1,d) = \sigma_1(m_1,d)$.
		\end{tabular}\end{center}
		\normalsize
	\item[Vector Concatenation:]
		Given two linguistic variables (with no apparent relationships)
		${\cal M}_0 \stackrel{\sigma_0}{\rightharpoondown} {\cal D}_0$
		and
		${\cal M}_1 \stackrel{\sigma_1}{\rightharpoondown} {\cal D}_1$,
		the {\em tensor product\/} is the linguistic variable
		\[ {\cal M}_0 \tensor {\cal M}_1 \stackrel{{\sigma_0}\tensor{\sigma_1}}{\rightharpoondown} {\cal D}_0 \tensor {\cal D}_1 \]
		on the tensor product space of terms and data,
		which products the term assignments
		\footnotesize
		\[ ({\sigma_0}\tensor{\sigma_1})((m_0,m_1),(d_0,d_1)) = \sigma_0(m_0,d_0) \tensor \sigma_1(m_1,d_1). \]
		\normalsize
\end{description}

We use the linguistic variable to interpret the meaning of 
the raw data assigned to objects by $\phi$.
This enriched interpretation,
called {\em granulation\/} in Fuzzy Sets 
or {\em conceptual scaling\/} in Formal Concept Analysis,
assigns a {\em view\/} or {\em facet\/} to the data $\phi$.
\begin{center}
	\begin{tabular}{|c|}\hline
		{\bf interpretation} \\ \hline\hline
		\mbox{\rule[-2mm]{0mm}{6mm}$\phi \stackrel{\sigma}{\longmapsto} \iota$} \\ \hline
	\end{tabular}
\end{center}
This facet takes the form of a {\bf V}-relation 
(an enriched formal context --- see below)
${\cal G} \stackrel{\iota}{\rightharpoondown} {\cal M}$
called the {\em derived context\/} in Formal Concept Analysis.
It is defined by relational composition
$\iota = \phi^\triangleright \circ \sigma^{\rm op}$.
In terms of elements this definition is
$\iota(g,m) 
	= \tilde{\sigma}(\phi(g))(m)
	= \sigma(m)(\phi(g))$.
The given indiscernibility $\gamma$ on objects $G$
is required to be as fine as 
the induced indiscernibility $\gamma_\phi$ on objects $G$,
defined via logical {\bf V}-equivalence
$\gamma_\phi(g_1,g_2) 
	= 
	\bigwedge_{m \in M} (\sigma(m)(\phi(g_1)) \Leftrightarrow \sigma(m)(\phi(g_2)))$.
Granulation of the tensor product of several linguistic variables
is called {\em apposition\/} in Formal Concept Analysis.

%The following commuting diagram of relations 
%illustrates our interpretation via linguistic variables.
%\begin{center}
%$\begin{array}{ccc}
%	{\cal G} &  \stackrel{\iota}{\rightharpoondown} & {\cal M} \\
%	\hspace{10pt}\mbox{\scriptsize$\phi$\normalsize} \searrow \hspace{-10pt} && \hspace{-10pt} \swarrow \mbox{\scriptsize$\sigma$\normalsize} \hspace{10pt} \\
%	& {\cal D} &
%\end{array}$
%\end{center}

%%%%%%%%%%%%%%%%%%%%%%%%%%%%%%%%%%%%%%%%%%%%%%%%%%%%%%%%%%%%%%%%%%%%%%%%%%%%%%%%
\section{Enriched Interpretation of Networked Information Resources}
%%%%%%%%%%%%%%%%%%%%%%%%%%%%%%%%%%%%%%%%%%%%%%%%%%%%%%%%%%%%%%%%%%%%%%%%%%%%%%%%

We are currently developing \cite{KeNe94}
an information management software system for the World-Wide Web
called {\sc wave},
the Web Analysis and Visualization Environment.
{\sc wave} is a third generation World-Wide Web tool 
used for navigation and discovery
over a universe of networked information resources.
Interpretation of resource descriptions,
via conceptual scaling or faceted analysis,
plays a central role in {\sc wave}.
At the present time,
the kernel component of the {\sc wave} system 
conceptually analyzes, interprets, and categorizes
resources,
such as Web textual and image documents,
in a crisp fashion.

However,
using ideas developed in this paper,
an excellent approach for the extension to an enriched {\sc wave} system
is quite clear.
The following short list of conceptually scalable attributes
indicates that notions of approximation are very important 
for networked information resources:
the visible size of textual documents in pages or some other meaningful unit;
the concept extent cardinality as a count of equivalent instances of resources;
similarity measures between Web documents based upon numbers of common attributes;
relative scores for {\sc waisindex} keyword search;
the cost of resources;
the duration of play for audio/video data;
the critical review of resources;
etc.
We intend to develop in the near future
an enriched {\sc wave} system,
which will allow the user to define according to his own judgement
various enriched interpretations of networked resource information.

%%%%%%%%%%%%%%%%%%%%%%%%%%%%%%%%%%%%%%%%%%%%%%%%%%%%%%%%%%%%%%%%%%%%%%%%%%%%%%%%
\appendix
%%%%%%%%%%%%%%%%%%%%%%%%%%%%%%%%%%%%%%%%%%%%%%%%%%%%%%%%%%%%%%%%%%%%%%%%%%%%%%%%

%%%%%%%%%%%%%%%%%%%%%%%%%%%%%%%%%%%%%%%%%%%%%%%%%%%%%%%%%%%%%%%%%%%%%%%%%%%%%%%%
\section{Closed Preorders}
%%%%%%%%%%%%%%%%%%%%%%%%%%%%%%%%%%%%%%%%%%%%%%%%%%%%%%%%%%%%%%%%%%%%%%%%%%%%%%%%
\label{closed_preorder}

A {\em closed preorder}
\cite{lawvere73}
${\bf V} = \quintuple{V}{\preceq}{\tensor}{\implication}{e}$
consist of the following data and axioms.
\begin{itemize}
	\item $\quadruple{V}{\preceq}{\tensor}{e}$ is a monoidal preorder,
		or ordered monoid,
		with $\pair{V}{\preceq}$ a preorder and $\triple{V}{\tensor}{e}$ a monoid,
		where the binary operation $\tensor \morph \product{V}{V} \rightarrow V$, called {\bf V}-composition, is monotonic:
		if both $u \preceq u'$ and $v \preceq v'$ then $(u \tensor v) \preceq (u' \tensor v')$.
	\item $\tensor$ is symmetric, or commutative; that is,
		$a \tensor b = b \tensor a$ for all elements $a,b \in V$.
	\item {\bf V} satisfies the closure axiom:
		the monotonic {\bf V}-composition function
		$(\:) \tensor b \morph V \rightarrow V$
		has a specified right adjoint
		$b \implication (\:) \morph V \rightarrow V$
		for each element $b \in B$,
		called {\bf V}-implication,
		or symbolically
		$\left( (\:) \tensor b \right) \dashv \left( b \implication (\:) \right) \morph V \rightarrow V$;
		that is,
		$a \tensor b \preceq c$ iff $a \preceq b \implication c$
		for any triple of elements $a,b,c \in V$.
	\item We usually also assume that our closed preorders are bicomplete;
		that is,
		the supremum $\bigvee B$ and the infimum $\bigwedge B$ exist
		(and are unique up to equivalence $\equiv$)
		for all subsets $B \subseteq V$.
\end{itemize}
The following define special closed preorders.
\begin{itemize}
	\item A closed preorder is {\em normal\/} 
		when the unit is the top element $e = \top_V$
		and {\bf V}-implication is directed-continuous:
		$b \implication(\bigvee_{d \in D}d) \equiv \bigvee_{d \in D}(b \implication d)$ for all directed subsets $D \subseteq V$.
		For normal closed preorders $a \tensor b \preceq a \wedge b$ for all elements $a,b \memberof V$.
	\item When the tensor product $\tensor$ is 
		the binary infimum or meet $\wedge$
		and the unit $e$ is the top element $\top_V$,
		the closed preorder
		${\bf V} = \quintuple{V}{\preceq}{\wedge}{\implication}{\top_V}$
		is called a {\em cartesian closed preorder\/}.
		The context of cartesian closed preorders is the context of traditional logic.
		A characteristic property of cartesian closed preorders is idempotency:
		$v \tensor v = v \wedge v = v$ for all elements $v \memberof V$.
		In a cartesian closed preorder,
		and even in an arbitrary closed preorder,
		we regard $V$ as being a set of generalized truth values.
		Cartesian closed preorders are normal.
\end{itemize}
We list some important closed preorders 
which can be used in Rough Sets and Soft Computing
for enriched interpretation in linguistic variables.
\begin{description}
	\item[Boolean truth-values] \mbox{ }
		\begin{center} 
			{\bf 2} = $\quintuple{2=\{0,1\}}{\leq}{\wedge}{\rightarrow}{1}$
		\end{center}
			where 0 is {\tt false}, 1 is {\tt true}, $\leq$ is the usual order on truth-values,
			$\wedge$ is the truth-table for {\tt and}, and $\rightarrow$ is the truth-table for {\tt implies}.
			Here {\bf 2}-spaces ${\cal X} = \pair{X}{d}$ are preorders ${\cal X} = \pair{X}{\preceq}$
			where $x_1 \preceq x_2$ when $d(x_1,x_2) = 1$,
			strict {\bf 2}-spaces are posets, and
			{\bf 2}-morphisms are monotonic functions.
	\item[Subset truth-values] \mbox{ }
		\begin{center} 
			\boldmath$\wp(A)$\unboldmath = $\quintuple{P(A)}{\subseteq}{\cap}{\rightarrow}{A}$
		\end{center}
			for any set $A$,
			where $P(A)$ is the set $P(A)=\{B \mid B \subseteq A\}$ of all subsets of $A$,
			$\cap$ is {\tt set intersection}, and $\rightarrow$ is {\tt set implication}:
			$B_1 \rightarrow B_2 = \{a \memberof A \mid a \memberof B_1 \mbox{ implies } a \memberof B_2\} = -B_1 \cup B_2$.
			\boldmath$\wp(A)$\unboldmath is essentially the marking space closed preorder 
			$\mbox{\boldmath$\wp(A)$\unboldmath} \cong {\bf 2}^A$
			defining the most basic markings-as-fuzzy-subsets interpretation for Petri nets.
	\item[Fuzzy truth-values] \mbox{ }
		\begin{center} 
			{\bf [0,1]} = $\quintuple{[0,1]}{\leq}{\wedge}{\rightarrow}{1}$
		\end{center}
			where 0 is {\tt false}, 1 is {\tt true}, 
			$0 \leq r \leq 1$ is some grade of truth-value between {\tt false} and {\tt true},
			$\leq$ is the usual order on fuzzy truth-values in the interval,
			$\wedge$ is the minimum operation representing the interval truth-table 
			for the fuzzy {\tt and},
			and $\rightarrow$ is operation
			$r \rightarrow s \define \left\{
			\begin{array}{cl}
				1,	&	r \leq s \\
				s,	&	r > s
			\end{array}
			\right.$
			representing the interval truth-table for the fuzzy {\tt implies}.
			The cartesian closed interval {\bf [0,1]} is coreflective and normal.
			This defines the correct context for Fuzzy Set theory.
	\item[Real truth-values] \mbox{ }
		\begin{center} 
			\boldmath$\Re$\unboldmath = $\quintuple{\Re=[0,\infty]}{\geq}{+}{\minus}{0}$
		\end{center}
			where $\geq$ is the usual downward ordering on the nonegative real numbers $\Re$
			(regarded as quantitative truth-values),
			$+$ is sum, 
			and $\minus$ defined by
			$s \minus r \define \left\{
			\begin{array}{cl}
				0,		&	r \geq s \\
				s - r,	&	r < s
			\end{array}
			\right.$
			representing the truth-table for the metrical {\tt difference}.
			The quantitative closed preorder of reals \mbox{\boldmath$\Re$\unboldmath} is normal.
\end{description}

\end{document}